\documentclass[sigconf,screen]{acmart}

\usepackage{graphicx}

\usepackage{amssymb}
\usepackage{multirow}
\usepackage{makecell}
\usepackage{booktabs}
\usepackage{enumitem}

\usepackage{balance}
\graphicspath{{image/}}

\AtBeginDocument{%
  \providecommand\BibTeX{{%
    \normalfont B\kern-0.5em{\scshape i\kern-0.25em b}\kern-0.8em\TeX}}}

\setcopyright{acmcopyright}
\copyrightyear{2022}
\acmYear{2022}
\setcopyright{acmcopyright}
\acmConference[MM '22] {Proceedings of the 30th ACM International Conference on Multimedia }{October 10--14, 2022}{Lisbon, Portugal.}
\acmBooktitle{Proceedings of the 30th ACM International Conference on Multimedia (MM '22), October 10--14, 2022, Lisbon, Portugal}
\acmPrice{15.00}
\acmISBN{978-1-4503-9203-7/22/10}
\acmDOI{10.1145/XXXXXX.XXXXXX}

\begin{document}
\title{Unbiased Directed Object Attention Graph for Object Navigation}

\author{Ronghao Dang}
\email{dangronghao@tongji.edu.cn}
\affiliation{%
  \institution{Tongji University}
  \country{}
}

\author{Zhuofan Shi}
\email{zhuofanshi@student.ethz.ch}
\affiliation{%
  \institution{Eidgenössische Technische Hochschule Zürich}
  \country{}
}

\author{Liuyi Wang}
\email{wly@tongji.edu.cn}
\affiliation{%
  \institution{Tongji University}
  \country{}
}

\author{Zongtao He}
\email{xingchen327@tongji.edu.cn}
\affiliation{%
  \institution{Tongji University}
  \country{}
}

\author{Chengju Liu}
\authornotemark[1]\thanks{*Corresponding author} 

\email{liuchengju@tongji.edu.cn}
\affiliation{%
  \institution{Tongji University}
  \country{}
}

\author{Qijun Chen}
\email{qjchen@tongji.edu.cn}
\affiliation{%
  \institution{Tongji University}
  \country{}
}
\renewcommand{\shortauthors}{Ronghao Dang, et al.}

\begin{abstract}

Object navigation tasks require agents to locate specific objects in unknown environments based on visual information. Previously, graph convolutions were used to implicitly explore the relationships between objects. However, due to differences in visibility among objects, it is easy to generate biases in object attention. Thus, in this paper, we propose a directed object attention (DOA) graph to guide the agent in explicitly learning the attention relationships between objects, thereby reducing the object attention bias. In particular, we use the DOA graph to perform unbiased adaptive object attention (UAOA) on the object features and unbiased adaptive image attention (UAIA) on the raw images, respectively. To distinguish features in different branches, a concise adaptive branch energy distribution (ABED) method is proposed. We assess our methods on the AI2-Thor dataset. Compared with the state-of-the-art (SOTA) method, our method reports 7.4\%, 8.1\% and 17.6\% increase in success rate (SR), success weighted by path length (SPL) and success weighted by action efficiency (SAE), respectively.
  
\end{abstract}





\maketitle
\begin{figure*}[t]
    \centering
    \includegraphics[width=\textwidth]{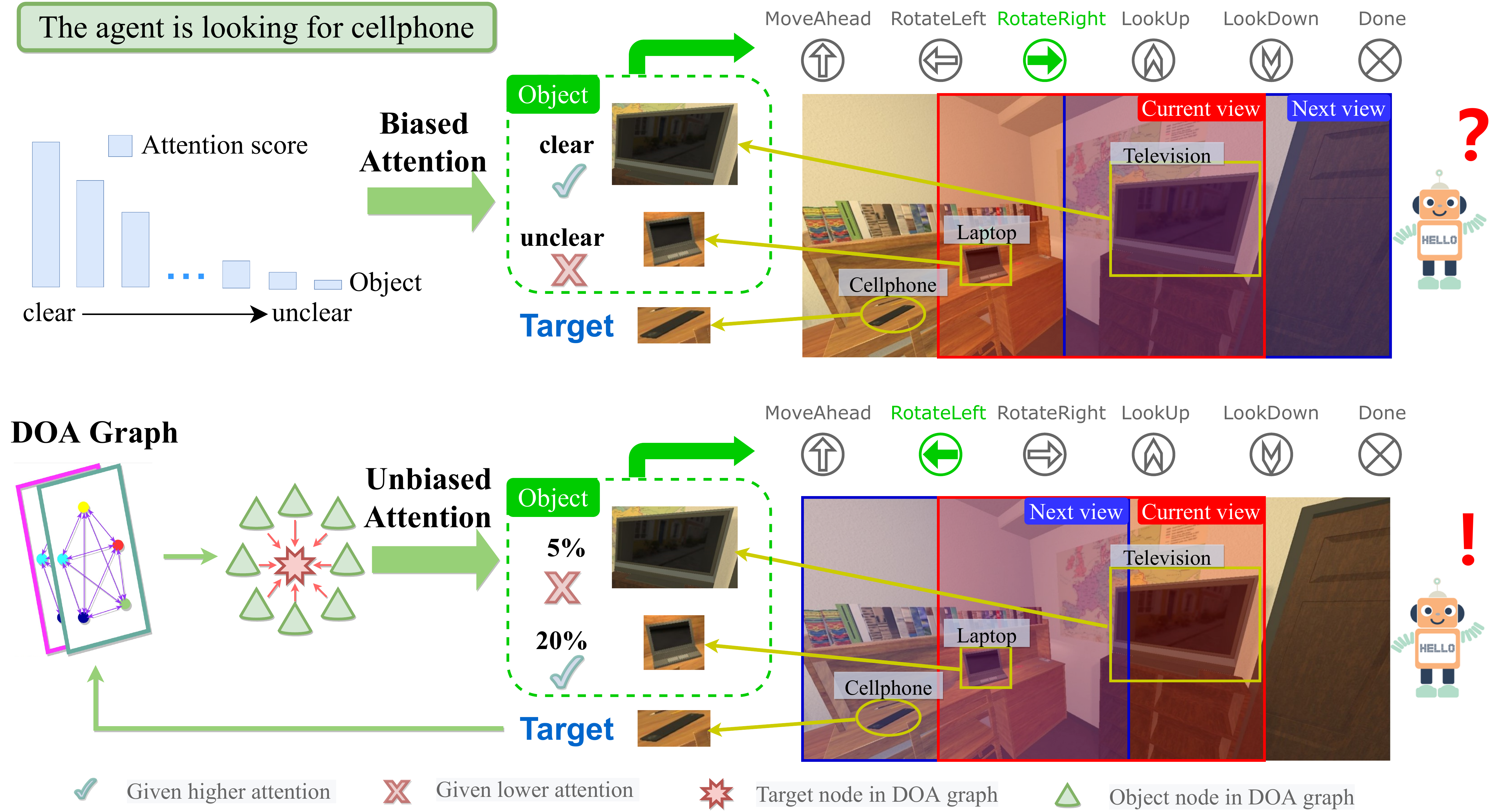}
    \caption{In the previous biased method \cite{du2020learning}, since the nearby TV is clearer than the distant laptop, the agent focuses more on the TV, resulting in an incorrect decision. Our proposed unbiased DOA graph learns that the cell phone is more likely to be near the laptop according to the target and the current view, resulting in an correct decision.}
    \label{fig:object attention bias}
\end{figure*}

\section{Introduction}

The object navigation \cite{ye2021auxiliary,moudgil2021soat,li2021ion,moghaddam2022foresi} requires an agent to navigate in  unseen environments and find a specified target by executing a sequence of actions. The agent can only use visual observation and target information as inputs and predict actions by deep reinforcement learning in each step.

Most prior works \cite{mirowski2016learning, mnih2016asynchronous, mnih2015human} have directly used global image features to recursively train agents based on egocentric observations. Nevertheless, if the target is invisible, it is difficult to efficiently navigate to the object position with these methods. Therefore, some recent works \cite{yang2018visual, du2020learning} have focused on establishing specific object prior knowledge to better understand complete scenes. Yang et al. \cite{yang2018visual} and Du et al. \cite{du2020learning} used the graph convolutional networks (GCNs) to learn graph representations of object features. 

However, the agent may not treat all items equally, preferring to focus on more conspicuous objects. Therefore, we propose the object attention bias problem, which refers to the situations in which agents focus on objects with high visibility during navigation. For example, when looking for a cell phone, an agent may focus on a closer, clearer TV while ignoring a distant, blurrier laptop that has a higher correlation with the cell phone (Figure~\ref{fig:object attention bias}). In principle, this phenomenon occurs mainly because neural networks prefer features with higher information entropy. Therefore, the direct use of graph convolutions to implicitly learn relationships between objects can lead to considerable object attention bias. 

To address the above problem, object attention should be decoupled from the object features, so we let the agent explicitly learn the attention weights of objects according to the different targets. Concretely, we propose a learnable two-layer directed object attention (DOA) graph that represents the relationships between objects in view and the target to address the object attention bias problem. The first graph layer is an intrinsic attention graph (IG) that establishes the basic object attention relationships. The second graph layer is a view adaptive attention graph (VAG) that changes based on the observations of the agent. The DOA graph, which is produced by the weighted average of the two graph layers, can guide an agent to reasonably assign attention to each object in view.

Based on the DOA graph, we further design two novel cross-attention modules: the unbiased adaptive image attention (UAIA) module and the unbiased adaptive object attention (UAOA) module. As illustrated in Figure~\ref{fig:object attention bias}, the target is the arrival point, and the objects in view are the starting points for the directed edges in the DOA graph. The weight of a directed edge from an object node to a target node is the object's attention while searching for this target. The UAOA module uses object attention in the DOA graph to directly distribute weights to different objects. The UAIA module uses a multihead attention mechanism to determine the areas in the global image that need attention. We follow the operation described in \cite{du2020learning} to concatenate the image branch, object branch and previous action branch into a vector. However, just as the temporal sequence in the transformer needs positional encodings to acquire position information \cite{vaswani2017attention,zhou2021informer,su2021roformer}, different branches need tokens to represent their identities. In consequence, we propose an adaptive branch energy distribution (ABED) method, which allows the network to distinguish different branches with the addition of few parameters. Then, in accordance with \cite{zhang2021hierarchical}, we input the concatenated features of the multimodel information into a long short-term memory (LSTM) network for learning and optimize the model with the A3C reinforcement learning strategy.

Extensive experiments on the AI2-Thor \cite{Kolve2017AI2THORAn} dataset  show that our method not only eliminates the object attention bias problem but also increases the state-of-the-art (SOTA) method by 7.4\%, 8.1\%, 17.6\% in the success rate (SR), success weighted by path length (SPL) and success weighted by action efficiency (SAE) \cite{zhang2021hierarchical}. Our method performs well inasmuch as the agent has a more comprehensive understanding of object relationships and an unbiased attention distribution. In summary, our contributions are as follows: 
\begin{itemize}[leftmargin=12pt, topsep=4pt, parsep=2pt]
\item We identify the prevalent object attention bias problem in object navigation, which occurs due to differences in object visibility.
\item We propose the directed object attention (DOA) graph, which addresses the problem of object attention bias and provides the agent with a better understanding of the internal relationships between objects.
\item The unbiased adaptive object attention (UAOA) and unbiased adaptive image attention (UAIA) modules use the DOA graph to allocate more reasonable attention resources to object features and global image features.
\item We propose the parameter-free adaptive branch energy distribution (ABED) method to optimize branch feature aggregation. 
\end{itemize}

\section{Related Works}



\subsection{Object Navigation}

In an object navigation task, an agent is given goal-oriented instruction to search for a specified object. The primitive object navigation models make decisions by directly processing input images with convolutional neural networks (CNNs).
Recently, researchers \cite{yang2018visual,gao2021room, chaplot2020object} have found that using only CNN features in raw images cannot achieve the desired results. An increasing number of researchers have begun to use methods such as object detection to extract high-level semantic features to better guide the agent's movement. Yang et al. \cite{yang2018visual} initially use  graph convolutional networks (GCNs) to learn the object prior knowledge.  Gao et al. \cite{gao2021room} utilize cross-modality knowledge reasoning (CKR) to apply an external knowledge graph in the agent's navigation. Zhang et al. \cite{zhang2021hierarchical} propose the hierarchical object-to-zone (HOZ) graph to guide an agent in a coarse-to-fine manner. In our work, we conduct the online-learning directed object attention (DOA) graph to serve as prior knowledge, which provides more unbiased object attention.

\subsection{Debiasing Methods}

Bias problems are widespread in machine learning \cite{torralba2011unbiased,kh2012undoing}, especially in the field of scene understanding \cite{tang2020unbiased, li2021bipartite}. However, no previous work has analyzed and addressed the bias problem in object navigation tasks. Current debiasing methods can be roughly categorized into three types: (i) data augmentation or re-sampling \cite{geirhos2018imagenet, li2018resound, li2019repair}, (ii) unbiased learning through the design of training networks and loss functions \cite{ zemel2013learning, lin2017focal}, and (iii) disentangling biased representations from unbiased representations \cite{misra2016seeing, cadene2019rubi}. Our proposed DOA graph method belongs to the second category. However, unlike common debiasing methods \cite{ zemel2013learning, lin2017focal}, our method explicitly models the bias module, which essentially solves the  object attention bias problem in object navigation. 

\section{Object Attention Bias}

\begin{figure}[t]
    \centering
    \includegraphics[width=0.45\textwidth]{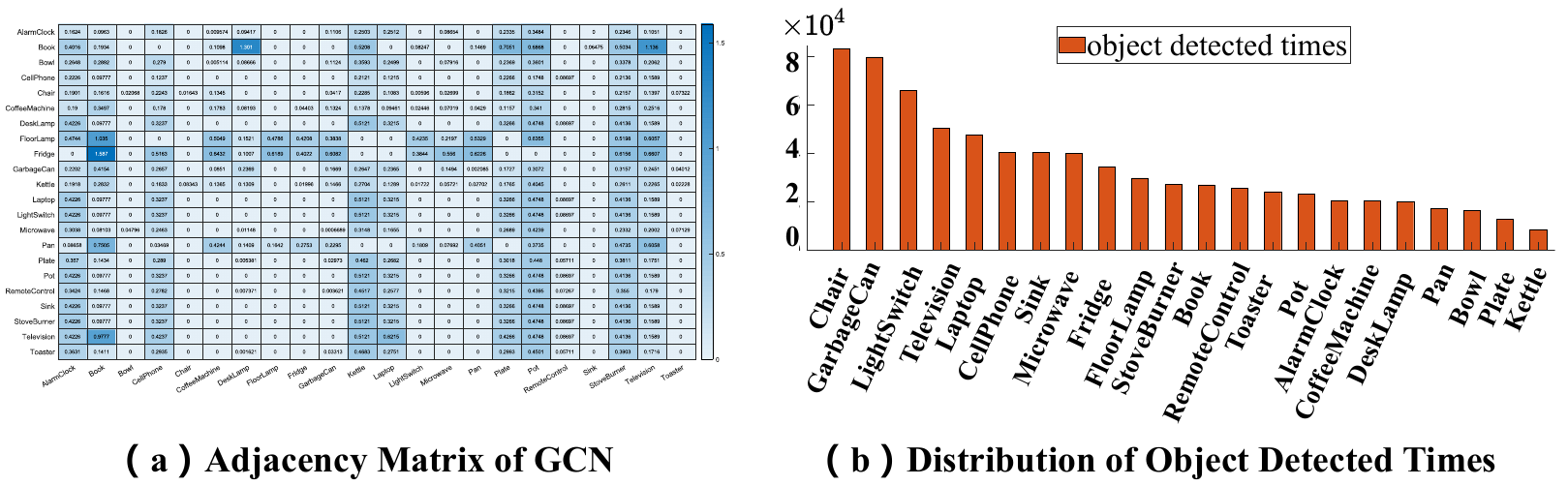}
    \caption{Two potential reasons for object attention bias: (a) endogenous cause: unreasonable GCN adjacency matrix \cite{du2020learning}; (b) exogenous cause: the total number of times each object is identified shows a long-tailed distribution.}
    \label{fig:four images}
\end{figure}

\subsection{Bias Discovered in GCNs}

The object GCN used in \cite{du2020learning, zhang2021hierarchical} attempts to aggregate the extracted object features by using the correlations between their bounding boxes. However, it is too difficult to learn a reasonable adjacency matrix which is crucial for GCNs \cite{zhou2020weighted}. As shown in Figure~\ref{fig:four images} (a), objects that are easy to observe, such as the floor lamp and fridge, have larger weights, while objects that are difficult to observe, such as the alarm clock and cell phone, have smaller weights. This kind of object attention bias is caused by over-focusing on coordinates and confidence scores. To reduce this bias, the agent should focus on what and where the object is rather than its size and clarity. The GCN ablation experiment, shown in Table~\ref{tab:branch ablation}, demonstrates that the GCN module only slightly improves the navigation ability of the agent, implying that a biased GCN module cannot be used to effectively and unbiasedly model relationships among objects.

\subsection{Duality of Bias}

We cannot criticize biased training because our visual world is inherently biased; people simply prefer to trust objects that are easier to identify. For example, when looking for a plate, we often pre-search for a cabinet instead of a knife or fork. In fact, some biased decisions allow agents to avoid some weird mistakes and make the overall actions more robust. However, excessive bias can cause an agent to overfit the training dataset, resulting in the agent ignoring critical navigation information. In general, there are two reasons for object attention bias: (\romannumeral1) endogenous cause (Figure~\ref{fig:four images} (a)), the network's own preference towards objects with richer visual features; (\romannumeral2) exogenous cause  (Figure~\ref{fig:four images} (b)), inequalities in the frequency each object are present in the dataset. This paper mainly starts from the endogenous cause without changing the number of objects in the dataset. Our proposed DOA graph corrects the agent's neglect of small and ambiguous objects (bad bias) while maintaining the agent's trust in high-confidence objects (good bias).

\section{Proposed Method}

\begin{figure*}[t]
    \centering
    \includegraphics[width=\textwidth]{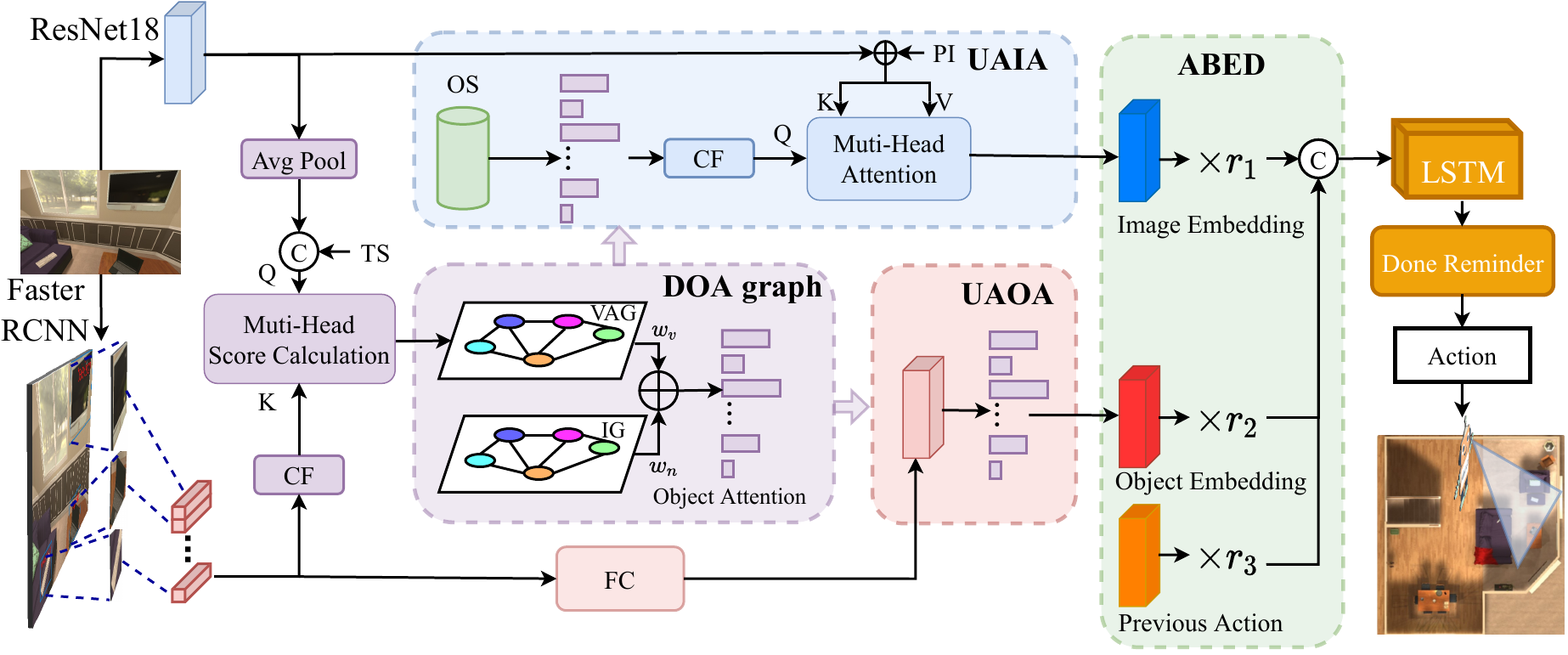}
    \caption{Model overview. PI: pixel index embedding, TS: target semantics, OS: object semantics, CF: confidence filter. VAG: view adaptive graph, IG: intrinsic graph, Avg Pool: average pooling. Our model consists of three branches: Image branch, Object branch, and Action branch. We perform UAIA on Image branch and UAOA on Object branch, respectively, based on directed object attention (DOA) graph. The joint features of the re-integrated branches after adaptive branch energy distribution (ABED) are input into an LSTM network to predict the next action.}
    \label{fig:model architecture}
\end{figure*}

Our goal is to propose an attention allocation strategy without object attention bias and a reasonable cross-branch aggregation method for a target-driven visual navigation system. To achieve this goal, our navigation system contains four major components, as illustrated in Figure~\ref{fig:model architecture}: 
(\romannumeral1) directed object attention (DOA) graph generator;
(\romannumeral2) unbiased adaptive object attention (UAOA) module; (\romannumeral3) unbiased adaptive image attention (UAIA) module; (\romannumeral4) adaptive branch energy distribution (ABED) method. (\romannumeral2) and (\romannumeral3) are based on the object attention in the DOA graph.

\subsection{Task Definition}

Initially, the agent is given a random target from a group of $N$ objects $P=\{Pan, \cdots,Cellphone\}$, and starts from a random state $ s=\{x,y,\theta _r,\theta _h\}$ in a random house. Here, $(x, y)$ and $(\theta _r, \theta _h)$ represent the coordinates and angles of the agent. After the state and target are initialized, the agent begins to navigate based on its own observations. At each timestamp $t$, the agent only receives the RGB image $o_t$ from a single perspective and the target $p\in P$.  According to $o_t$ and $p$, the agent learns a navigation strategy $\pi(a_t|o_t, p)$, where $a_t\in A=\{MoveAhead$; $RotateLeft$; $RotateRight$;  $LookDown$;  $LookUp$; $Done\}$ and $Done$ is the output if the agent believes it has navigated to the target location. The successful episode is defined as: an agent selects the termination action $Done$ when the distance between the agent and the target is less than 1.5 meters and the target is in the field of view.

\subsection{Directed Object Attention (DOA) Graph}

DOA graph is a graphical representation of the correlation degree between identifiable objects and the target. According to the analysis presented in Section 3.1, previous GCN-based methods cannot learn unbiased relationships between objects in object navigation tasks. In contrast, our proposed DOA graph provides an explicit yet flexible representation of the relationships between objects. The DOA graph is obtained through a weighted summation of the intrinsic object graph and the view adaptive graph.

\subsubsection{Intrinsic Object Graph}

The intrinsic object graph represents the ubiquitous intrinsic relationships between objects. For example, a mouse and a laptop have a strong inherent relationship. Here, we define a learnable matrix $ G_n \in \mathbb{R}^{N \times N}$ to represent the intrinsic object graph, where $N$ is the number of objects. As the agent uses reinforcement learning to continuously explore different environments, $G_n$ gradually tends to be reasonable and stable. $G_n$ is fixed during testing. Each edge between objects in $G_n$ is bidirectional. The end node of an edge represents the target object $p$, while the start node of the edge represents the object that needs to be assigned attention. Therefore, the weight of each directed edge represents the intrinsic correlation between an object  $q \in P$ and the target object $p \in P$. Each row of $G_n$ is normalized using SoftMax to ensure that the sum of all edge values connected to a target node is 1.

\subsubsection{View Adaptive Graph}

The view adaptive graph represents the real-time adaptive relationships between objects. The agent generates the global image features $ I_t \in \mathbb{R}^{M \times 512} $ (from ResNet18 \cite{he2016deep}) and object features $ S_t \in \mathbb{R}^{N \times 518}$ (from Faster-RCNN \cite{ren2015faster}) after observing the current scene. Here, $M$ is the pixel number of the image feature map. $S_t$ is concatenated by the object visual features $ S_t^{visual}$, object position features $ S_t^{pos}$, confidence $S_t^{conf}$ and target indicator $S_t ^{target}$. Since low-confidence object detection results often contain excessive noise, we filter $S_t$ with the confidence criterion $S_t^{conf}>threshold$ to obtain $\widetilde{S_t}$. 

For introducing target information to the image features $I_t$, we encode the object index using the one-hot method and two fully connected layers to obtain $OI$. The input image query $\widetilde{I_t} \in \mathbb{R}^{1 \times 576}$ can be formulated as:
\begin{equation}
\widetilde{I_t}= Concat(\frac{1}{M}\sum_{i=1}^{M}I_{t}^{i},OI^p)
\end{equation}
where $OI^p$ refers to the $p$-th object (target) semantics. $\frac{1}{M}\sum_{i=1}^{M}I_{t}^{i}$ is squeezing global spatial information into a channel descriptor  using global average pooling. The agent grounds the current overall environmental characteristics  $\widetilde{I_t}$ to the object features $\widetilde{S_t}$ via muti-head score calculation \cite{vaswani2017attention} to produce the view adaptive graph $G_v \in \mathbb{R}^{N \times 1}$:
\begin{gather}
\widetilde{Q}_i = \widetilde{I_t}\widetilde{W}_i^Q \;\;\;\;\;  \widetilde{K}_i=\widetilde{S_t}\widetilde{W}_i^K \;\;\;\;\; 
i=1,2,\cdots ,NH\\
\widetilde{head}_i=softmax(\frac{\widetilde{Q}_i\widetilde{K}_i^T}{\sqrt{HD}}) \\
G_v = Concat(\widetilde{head}_1,\cdots ,\widetilde{head}_{NH})\widetilde{W}^O
\end{gather}
where $HD$ and $NH$ denote the hidden dimensionality and number of heads in the muti-head score calculation. $\widetilde{W}_{i}^{Q} \in \mathbb{R}^{576 \times HD}$ and $\widetilde {W}_{i}^{K} \in \mathbb{R}^ {518 \times HD}$ map $\widetilde{I_t}$ and $\widetilde{S_t}$ to the same dimension $HD$. $\widetilde{W}^O \in \mathbb{R}^{NH \times 1}$ aggregate the various subgraphs calculated by the scaled dot-product of the multiple heads to generate the graph $G_v$.

\begin{table*}[t]
\setlength\tabcolsep{4.8pt}
\caption{We obtain a strong and concise baseline by comparing the roles of different modules in previous methods \cite{du2020learning, zhang2021hierarchical, qiu2020target} (\%). These modules include the object GCN (GCN), the row image branch (Image), the zone branch (Zone), the room branch (Room), the previous action branch (Action) and the object branch (Object).}
\label{tab:branch ablation}
\begin{tabular}{cccccc|ccc|ccc|c}
\hline
\multirow{2}*{GCN} & \multirow{2}*{Image} & \multirow{2}*{Zone} & \multirow{2}*{Room} & \multirow{2}*{Action} & \multirow{2}*{Object} & \multicolumn{3}{c|}{All} & \multicolumn{3}{c|}{L\textgreater{}=5} & \multirow{2}*{\makecell[c]{Episode \\ Time}} \\
 &  &  &  &  &  & SR & SPL & SAE & SR & SPL & SAE &  \\ \hline
\checkmark & \checkmark & \checkmark & \checkmark & \checkmark & \checkmark & $68.53_{\pm1.32}$  & $37.50_{\pm1.01}$  & $25.98_{\pm0.96}$  & $60.27_{\pm1.67}$  & $36.61_{\pm1.12}$  & $27.68_{\pm1.31}$  & 0.239 \\ \hline
 & \checkmark & \checkmark & \checkmark & \checkmark & \checkmark & $68.11_{\pm1.61}$  & $37.80_{\pm0.87}$  & $26.07_{\pm1.26}$  & $58.15_{\pm1.19}$  & $36.30_{\pm0.97}$  & $27.55_{\pm0.88}$  & 0.177 \\
\checkmark &  & \checkmark & \checkmark & \checkmark & \checkmark & $63.95_{\pm1.37}$  & $32.04_{\pm1.69}$  & $26.72_{\pm0.87}$  & $54.38_{\pm0.56}$  & $30.38_{\pm1.27}$  & $26.52_{\pm1.21}$  & 0.145 \\
\checkmark & \checkmark &  & \checkmark & \checkmark & \checkmark & $67.51_{\pm0.97}$  & $37.81_{\pm1.13}$  & $25.75_{\pm1.04}$  & $58.31_{\pm1.22}$  & $36.69_{\pm0.87}$  & $27.83_{\pm1.36}$  & 0.223 \\
\checkmark & \checkmark & \checkmark &  & \checkmark & \checkmark & $67.61_{\pm1.16}$  & $37.60_{\pm1.32}$  & $26.11_{\pm1.88}$  & $59.21_{\pm1.45}$  & $36.62_{\pm0.99}$  & $28.35_{\pm1.17}$  & 0.235 \\
\checkmark & \checkmark & \checkmark & \checkmark &  & \checkmark & $64.46_{\pm1.74}$  & $34.60_{\pm1.14}$  & $25.49_{\pm1.09}$  & $54.68_{\pm0.67}$  & $33.01_{\pm0.93}$  & $25.81_{\pm1.12}$  & 0.243 \\
\checkmark & \checkmark & \checkmark & \checkmark & \checkmark &  & $48.26_{\pm1.11}$  & $19.88_{\pm1.31}$  & $17.58_{\pm1.57}$  & $35.80_{\pm1.12}$  & $17.23_{\pm1.65}$  & $16.66_{\pm1.18}$  & 0.160 \\ \hline
 & \checkmark &  &  & \checkmark & \checkmark & $69.14_{\pm0.67}$  & $37.87_{\pm0.88}$  & $27.77_{\pm0.95}$  & $60.42_{\pm1.11}$  & $37.28_{\pm0.66}$  & $28.70_{\pm0.67}$  & 0.174 \\ \hline
\end{tabular}
\end{table*}

\subsubsection{Object Attention} 

According to the searched target $p$, we take the edge weight $G_{n}^{p} \in \mathbb{R}^{N \times 1}$ from the intrinsic object graph $G_n$ with the $p$-th node as the end point. With the weighted summation of the intrinsic weight and the view adaptive weight, we can obtain the attention:
\begin{equation}
    G_t = G_{n}^{p}w_n + G_{v}w_v
\end{equation}
that each object requires. The learnable $w_n$ and $w_v$ are the weights of the two graphs.

\subsection{Unbiased Adaptive Attention}

\subsubsection{Unbiased Adaptive Object Attention (UAOA)}

The purpose of unbiased adaptive object attention (UAOA) is to use the object attention $G_t$ obtained in Section 4.2 to assign attention to different object features. To balance the information integrity and computational complexity, we apply two fully connected layers around the ReLU \cite{nair2010rectified} to map the object features $S_t$ to a lower-dimensional representation
${S_{t}}' \in \mathbb{R}^{N \times 64}$.
Finally, the attention weight of each object $q$ is multiplied by the object features ${S_{t}}’$:
\begin{equation}
 \widehat{S_{t}^{q}} = F_{scale}( {S_{t}^{q}}’, G_{t}^{q}) = {S_{t}^{q}}’ G_{t}^{q} \;\;\;\;\;\;\;\;
 q = 1,2,\cdots, N
\end{equation}
where ${S_{t}}’=\{{S_{t}^{1}}',{S_{t}^{2}}',\cdots ,{S_{t}^{N}}'\}$,  ${S_{t}^{q}}’$ is the low dimensional feature of the $q$-th object. $G_{t}^{q}$ is the weight of the $q$-th object in DOA graph at time $t$. 

\subsubsection{Unbiased Adaptive Image Attention (UAIA)}

We use the DOA graph to focus more attention on the areas around important objects in the image. We use the encoded object index $OI$ rather than word embeddings trained by other networks to represent the object semantic information (what the object is). The object index embeddings learned by our network are more coupled to our dataset than word embeddings trained on other tasks. We use the object attention $G_{t}$ to generate the attention-aware object semantics $D$:
\begin{equation}
D=\sum_{q=1}^N\mathbb{I}(S_t^{conf}>threshold) G_{t}^{q}\times{OI^{q}}
\end{equation}
where $\mathbb{I}(\cdot )$ is the indicator function. The confidence filter $S_t^{conf}>threshold$ indicates that only the semantics of objects whose detection confidence exceeds $threshold$ are used. Unlike the convolution operation in CNNs, the muti-head attention operation is permutation-invariant \cite{dosovitskiy2020image}, which cannot leverage the order of the tokens in an input sequence. To mitigate this issue, our work adds absolute positional encodings to each pixel in the input sequence, such as in \cite{vaswani2017attention, dosovitskiy2020image}, enabling order awareness. We use standard learned 1D pixel index embedding $PI \in \mathbb{R}^{M \times 64}$ since we have not observed any significant performance gains when using more complicated 2D position embeddings. The resulting sequence of embedded vectors serves as input to the encoder.
\begin{equation}
{I_t}'=\delta(\delta(I_{t}W_{1})W_{2})+PI
\end{equation}
where $\delta$ refers to the ReLU function, $W_1 \in \mathbb{R}^{512 \times 128} $ and $W_2 \in \mathbb{R}^{128 \times 64}$ reduce the dimension of the global image features $I_t$ and the pixel index embedding $PI$ is introduced to generate position-aware image features ${I_t}’ \in \mathbb{R}^{M \times 64}$. We use the attention-aware object semantics $D$ as the query and the position-aware image features ${I_t}'$ as the key and value in the muti-head image attention ($HD=64$, $NH=4$)  to generate the final image embedding $\widehat {I_{t}}$. 
\begin{gather}
Q_i = D W_i^Q \;\;\;\;
K_i= {I_t}’ W_i^K \;\;\;\;
V_i= {I_t}’ W_i^V \;\;\;\;
i=1,2,\cdots ,NH\\
head_i=softmax(\frac {Q_{i} K_{i}^T}{\sqrt{HD}})V_i \\
\widehat {I_{t}} = Concat(head_1,\cdots , head_{NH})W^O
\end{gather}

\subsection{Adaptive Branch Energy Distribution (ABED)}

Previous works \cite{zhang2021hierarchical, du2020learning} have input the directly concatenated embedded features of the three branches (object, image and previous action branches) into LSTM networks to establish time series models. However, there are two issues with this simple feature stitching: (\romannumeral1) It is difficult for the network to distinguish between the different information types of the three branches during training; (\romannumeral2) It is difficult to guarantee that the data distribution of the concatenated vector is rational. Therefore, we propose the adaptive branch energy distribution (ABED) method to provide additional tokens to each branch without introducing extra parameters, and optimize the data distribution of the concatenated vector. We establish a learnable vector $R=\{r_1,r_2,r_3\}$ with only three parameters. The final output vector $H_t$ can be expressed as:
\begin{equation}
    H_t=F_{pw}(Concat(r_1\widehat{S_t},r_2\widehat{I_t},r_3PA))
\end{equation}
where $PA$ is the previous action embedding and $F_{pw}$ refers to the pointwise convolution. The operation, which is similar to the energy distribution of the input signal, uses the significant differences between the feature distributions of the three branches to explicitly distinguish the branches and learn a more reasonable distribution of the concatenated vector.  Although our method is quite simple, experiments have proven that it can significantly improve the navigation ability of the agent. When compared to the process of directly adding branch semantic (BS) embeddings to each branch, this method is unique in that it can provide the model with a strong scene understanding without destroying the other modules in complex models.

\subsection{Policy Learning}

Previous works \cite{wortsman2019learning, yang2018visual} have used direct observations to learn a strategy $\pi(a_t|o_t, p)$. Our work uses unbiased object graph relationships to learn an LSTM action policy $\pi(a_t|H_t, p)$, where $H_t$ is the joint representation of the global image embedding $r_1 \widehat{I_{t}}$, object embedding $r_2\widehat{S_{t}}$ and previous action embedding $r_{3}PA$. Based on previous works \cite{ mirowski2016learning, fang2021target}, we treat this task as a reinforcement learning problem and utilize the asynchronous advantage actor-critic (A3C) algorithm \cite{ mnih2016asynchronous}, which applies policy gradients to assist the agent in choosing an appropriate action $a_t$ in the high-dimensional action space $A$. In accordance with the done reminder operation presented in \cite{zhang2021hierarchical}, we use the target detection confidence to explicitly enhance the probability of the $Done$ action.

\section{Experiments}

\begin{table}[t]
\caption{The ablation study of the different components of the UAIA module (\%). CF: confidence filter, PI: pixel index.}
\label{tab:UAIA}
\begin{tabular}{cc|ccc|ccc}
\hline
\multicolumn{2}{c|}{UAIA} & \multicolumn{3}{c|}{ALL} & \multicolumn{3}{c}{L>=5} \\
CF & PI & SR & SPL & SAE & SR & SPL & SAE \\ \hline \hline
 &  & 70.48 & 37.13 & 27.63 & 63.12 & 37.09 & 29.16 \\ \hline
\checkmark &  & 72.12 & 37.45 & 27.98 & 64.23 & 37.67 & 29.13 \\
 & \checkmark & 71.12 & 38.91 & 27.23 & 64.01 & 39.34 & \textbf{30.33} \\ \hline
\checkmark & \checkmark & \textbf{72.28} & \textbf{39.13} & \textbf{28.91} & \textbf{65.19} & \textbf{39.98} & 30.22 \\ \hline
\end{tabular}
\end{table}

\begin{table}[t]
\setlength\tabcolsep{4.5pt}
\caption{The ablation study of the different modules in the DOA graph (\%). IG: intrinsic graph, VAG: view adaptive graph, MA: multi-head attention, TS: target semantics. '------' indicates that the view adaptive graph is not used.}
\label{tab:DOA}
\begin{tabular}{ccc|ccc|ccc}
\hline
\multirow{2}{*}{IG} & \multicolumn{2}{c|}{VAG} & \multicolumn{3}{c|}{ALL} & \multicolumn{3}{c}{L\textgreater{}=5} \\
 & MA & TS & SR & SPL & SAE & SR & SPL & SAE \\ \hline \hline
 &  &  & 70.77 & 38.51 & 27.98 & 63.83 & 38.04 & 28.74 \\ \hline
\checkmark & \multicolumn{2}{c|}{\textbf{------}} & 73.67 & 39.98 & \textbf{30.23} & 65.79 & 39.23 & 31.98 \\
\checkmark &  &  & 72.12 & 39.29 & 28.10 & 65.55 & 39.21 & 30.91 \\
\checkmark & \checkmark &  & 73.91 & 39.21 & 30.21 & 67.78 & 39.34 & 31.89 \\
 & \checkmark & \checkmark & 67.43 & 36.29 & 27.01 & 60.12 & 35.90 & 26.88 \\ \hline
\checkmark & \checkmark & \checkmark & \textbf{74.32} & \textbf{40.27} & 29.79 & \textbf{67.88} & \textbf{40.36} & \textbf{32.56} \\ \hline
\end{tabular}
\end{table}

\subsection{Experimental Setup}

\subsubsection{Dataset}

We choose the AI2-Thor \cite{ Kolve2017AI2THORAn} dataset and its corresponding simulator as the experimental platform. The AI2-Thor dataset includes 30 different floorplans for each of 4 room layouts: kitchen, living room, bedroom, and bathroom. For each scene type, we use 20 rooms for training, 5 rooms for validation, and 5 rooms for testing. There are 22 kinds of objects ($N=22$) that agents can recognize, and we ensure that there are at least 4 kinds of objects in each room \cite{ wortsman2019learning}.

\subsubsection{Evaluation Metrics}

We use the success rate (SR), success weighted by path length (SPL) \cite{ anderson2018evaluation}, and success weighted by action efficiency (SAE) \cite{ zhang2021hierarchical} metrics to evaluate our method. SR indicates the success rate of the agent in completing the task, which is formulated as $SR = \frac{1}{E}\sum_{i=1}^{E}Suc_i$, where $E$ is the number of episodes and $Suc_i$ indicates whether the $i$-th episode succeeds. SPL considers the path length more comprehensively and is defined as $SPL = \frac{1}{E}\sum_{i=1}^{E}Suc_{i}\frac{L_i^*}{max(L_i, L_i^*)}$, where $L_i$ is the path length taken by the agent and $L_i^*$ is the theoretical shortest path. SAE considers the effects of unnecessary rotations and is defined as $ SAE = \frac{1}{E}\sum_{i=1}^{E}Suc_i\frac{\sum_{t=0}^{T}\mathbb{I}(a_t^i \in A_{change})}{\sum_{t=0}^{T}\mathbb{I}(a_t^i \in A_{all})}$, where $\mathbb{I}(\cdot)$ is the indicator function, $a_t^i$ is the agent’s action at time $t$ in episode $i$, $A_{all}$ is the set of all actions, and $A_{change}$ is the set of actions that can change the position of the agent.

\begin{table}[t]
\setlength\tabcolsep{4.5pt}
\caption{The ablation study of different branch fusion methods in ABED (\%). CA: cross-attention, BS: branch samentics, ED: energy distribution method.}
\label{tab:ABED}
\begin{tabular}{c|cc|ccc|ccc}
\hline
\multirow{2}{*}{CA} & \multicolumn{2}{c|}{ABED} & \multicolumn{3}{c|}{ALL} & \multicolumn{3}{c}{L\textgreater{}=5} \\
 & BS & ED & SR & SPL & SAE & SR & SPL & SAE \\ \hline \hline
 &  &  & 69.14 & 37.87 & 27.77 & 60.42 & 37.28 & 28.70 \\ \cline{2-9} 
 & \checkmark &  & 71.07 & 40.02 & 26.30 & 64.04 & 39.50 & 29.57 \\
 &  & \checkmark & 71.56 & 38.66 & 28.99 & 64.12 & 38.89 & 29.07 \\ \hline \hline
\multirow{3}{*}{\checkmark} &  &  & 73.44 & 39.55 & 29.12 & 67.01 & 39.34 & 30.55 \\ \cline{2-9} 
 & \checkmark &  & 67.82 & 33.50 & 27.74 & 61.63 & 34.23 & 29.10 \\
 &  & \checkmark & \textbf{74.32} & \textbf{40.27} & \textbf{29.79} & \textbf{67.88} & \textbf{40.36} & \textbf{32.56} \\ \hline
\end{tabular}
\end{table}

\begin{table}[t]
\setlength\tabcolsep{3.7pt}
\small
\caption{The ablation study of the different components in overall model (\%).}
\label{tab:overall ablation}
\begin{tabular}{ccc|ccc|ccc}
\hline
\multirow{2}{*}{UAOA} & \multirow{2}{*}{UAIA} & \multirow{2}{*}{ABED} & \multicolumn{3}{c|}{ALL} & \multicolumn{3}{c}{L\textgreater{}=5} \\
 &  &  & SR & SPL & SAE & SR & SPL & SAE \\ \hline \hline
 &  &  & 69.14 & 37.87 & 27.77 & 60.42 & 37.28 & 28.70 \\ \hline
\checkmark &  &  & 73.21 & 39.20 & 29.29 & 66.37 & 39.12 & 31.18 \\
 & \checkmark &  & 72.28 & 39.13 & 28.91 & 65.19 & 39.98 & 30.22 \\
 &  & \checkmark & 70.77 & 38.51 & 27.98 & 63.83 & 38.04 & 28.74 \\ 
 \checkmark & \checkmark &  & 73.58 & 39.11 & 29.46 & 67.42 & 39.04 & 31.74 \\ \hline
\checkmark & \checkmark & \checkmark & \textbf{74.32} & \textbf{40.27} & \textbf{29.79} & \textbf{67.88} & \textbf{40.36} & \textbf{32.56} \\ \hline
\end{tabular}
\end{table}

\subsubsection{Implementation Details}

We train our model with 18 workers on 2 RTX 2080Ti Nvidia GPUs. The Adam optimizer \cite{ kingma2014adam} is used to update the network parameters with a learning rate of $10^{-4}$. We introduce a dropout of 0.3 to the muti-head attention mechanism and global image embedding. The confidence threshold for filtering objects  is set to 0.6. Faster-RCNN \cite{ ren2015faster} is fine-tuned on 50\% \cite{zhang2021hierarchical} of the training data from the AI2-Thor dataset. For evaluation, our results take the average value of the test for 3 times. We report the results for all targets (ALL) and for a subset of targets (L>=5) with optimal trajectory lengths greater than 5. 

\begin{figure*}[t]
    \centering
    \includegraphics[width=0.98\textwidth]{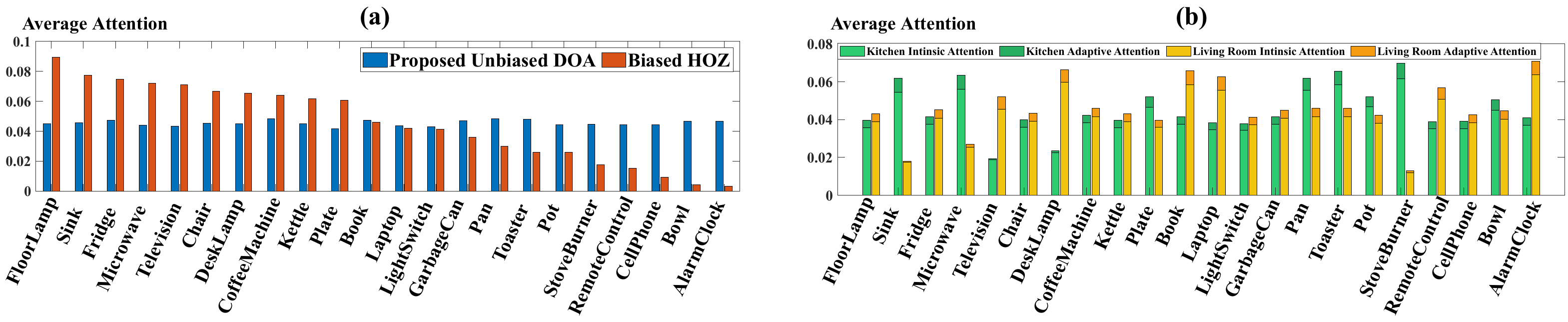}
    \caption{
    (a) Compare the object average attention of biased HOZ \cite{zhang2021hierarchical} and our unbiased DOA method.
    (b) Compare the object average attention (intrinsic attention + adaptive attention) of different scenes in DOA graph.
    }
    \label{fig:proposed unbias attention}
\end{figure*}

\subsection{Strong and Concise Baseline}

The methods presented in \cite{du2020learning, zhang2021hierarchical, qiu2020target} include five kinds of branches (image, zone, room, previous action and object) that are concatenated into a vector before being input into the LSTM network. To evaluate the influence of the five branches separately on the object navigation task, we eliminate each branch in turn, as shown in Table~\ref{tab:branch ablation}. The object branch has the greatest impact on the experimental results, and the removal of the object branch drops 20.27/24.27, 17.62/19.38, 8.40/11.02 in SR, SPL and SAE (ALL/L>=5, \%). This confirms the importance of object features in object navigation task. In this perspective, the adequate exploration of object relationships is necessary. Moreover, the image and previous action branches also have significant impacts on the agent's navigation ability. Whereas, the room branch and the zone branch have little effect on the SR and SPL. Accordingly, our baseline retains only the image, object and previous action branches. 

The last row in Table~\ref{tab:branch ablation} shows our simplified baseline. The removal of the object GCN, room branch and zone branch reduces the agent's exploration time by 27.2\% while leaving the SR, SPL and SAE essentially unchanged. This more concise baseline allows us to observe the advantages of the added modules more clearly.

\subsection{Ablation Experiments}

We verify the effectiveness of each proposed module with extensive experiments. Table~\ref{tab:UAIA}, Table~\ref{tab:DOA}, Table~\ref{tab:ABED} and Table~\ref{tab:overall ablation} show the results of ablation experiments on the UAIA, DOA graph, ABED and overall model. More ablation experiments are provided in the Supplementary Material. 

\subsubsection{UAIA}
As shown in Table~\ref{tab:UAIA}, the UAIA module has two main components: the confidence filter (CF),  which is used to eliminate outlier objects; the pixel index (PI) embedding, which is used to increase the order-awareness of the global image features. The UAIA module with CF (confidence > 0.6) outperforms the UAIA module without CF by 1.64/1.11 in SR (ALL/L >= 5, \%). This result shows that reducing the influence of irrelevant objects in the UAIA module can effectively improve the navigation ability. The UAIA module with PI embedding outperforms the UAIA module without PI embedding by 1.78/2.25 in SPL (ALL/L >= 5, \%), demonstrating that adding positional encoding to each pixel in the image can optimize the agent's path. The two components complement each other to improve the effect of the UAIA module.

\subsubsection{DOA Graph}
As shown in Table~\ref{tab:DOA}, we explore the role of intrinsic graph (IG) and view adaptive graph (VAG) in DOA graph. The DOA graph with only IG outperforms the DOA graph without IG by 2.90/1.96, 1.47/1.19 and 2.25/3.24 in SR, SPL and SAE (ALL/L >= 5, \%).  However, the DOA graph with only VAG brings a blow to the agent's navigation ability. This result implies that it is difficult to directly perform fully adaptive  learning in object relationships, requiring the extensive prior knowledge to narrow the learning domain. During the calculation of VAG, the muti-head attention (MA) allows for more reasonable object graph relationships, while the use of target semantics (TS) allows the agent to be more specific about the target, thereby improving the navigation efficiency. 

\subsubsection{ABED}
In Table~\ref{tab:ABED}, We compare two methods of providing identities for the branches: branch semantics (BS), which use the embedding of one-hot vectors; energy distribution (ED), which uses only three energy distribution coefficients. Without cross-attention (UAIA and UAOA), adding BS and ED to the original model improves the agent's navigation ability well. Notably, with cross-attention (UAIA and UAOA), adding BS to the complex model causes the model to crash. In contrast, the simple ED method improves the complex model with cross-attention by 0.88/0.87, 0.72/1.02 and 0.67/2.01 in SR, SPL and SAE (ALL/L >= 5, \%). The results demonstrate the significant advantage of the ED method with only three parameters in complex models. This is consistent with our intuition that due to the complexity of object navigation tasks, the learning model must be simplified; otherwise, the strong coupling of the complex parameters between modules can cause the overall model to be difficult to learn. 

\subsubsection{Overall Model} 
The ablation experiments on the overall model with UAOA, UAIA and ABED are shown in Table~\ref{tab:overall ablation}. Compared with our proposed baseline, applying the complete model increases SR, SPL and SAE by 5.18/7.46, 2.40/3.08 and 2.02/3.86 (ALL/L >= 5, \%). The results indicate that our method is capable of effectively guiding agents to navigate in unseen environments. Compared with the UAIA and ABED methods, the UAOA method improves the model  more significantly. This is because the UAOA method essentially solves the object attention bias problem, increasing the agent's understanding of the relationships between objects.

\begin{figure*}[t]
    \centering
    \includegraphics[width=0.8\textwidth]{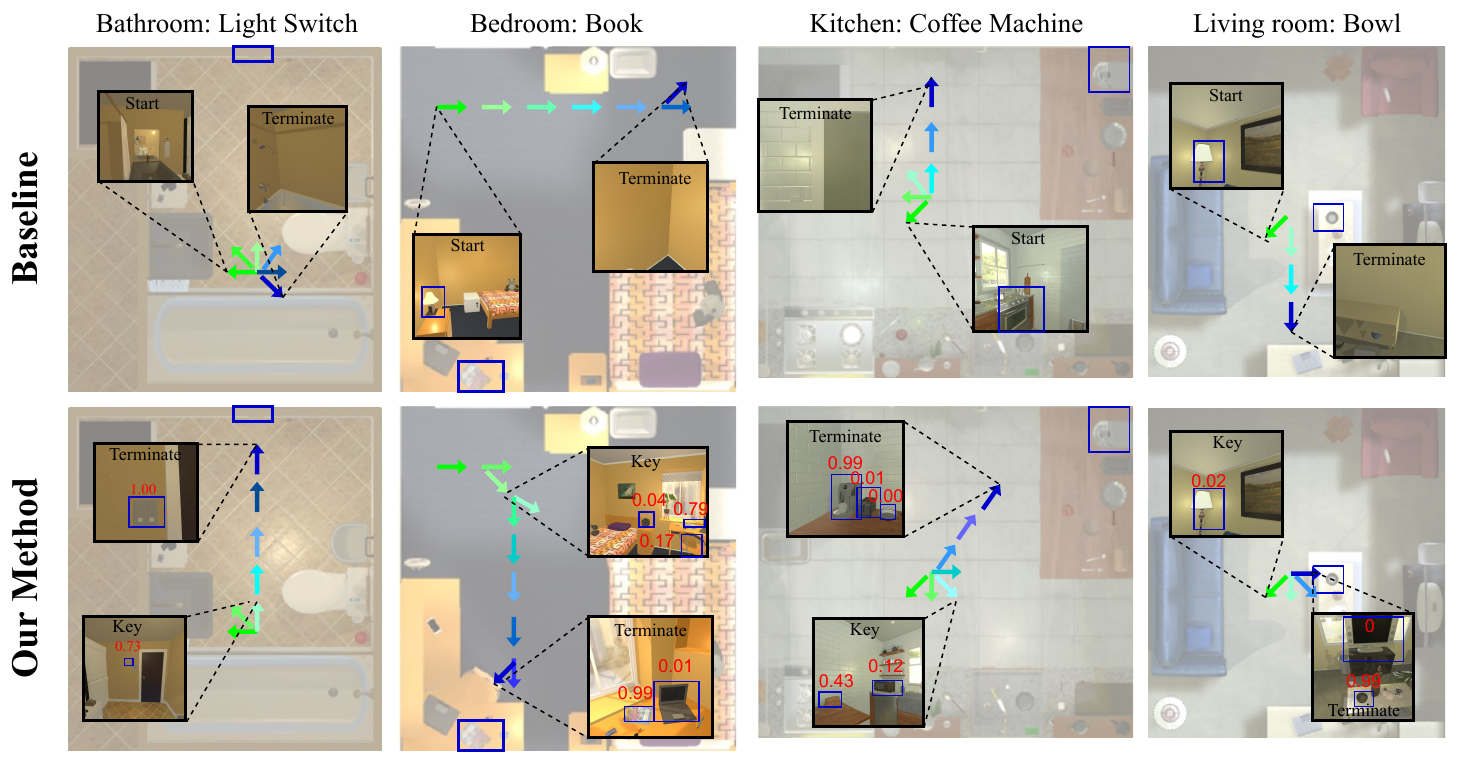}
    \caption{Visualization in testing environment. We show the results of experiments that involve searching for four different objects in four different scenes. The trajectory of the agent is indicated by green and blue arrows, where green is the beginning and blue is the end. The red value in the object detection box represents the attention weight of the object.}
    \label{fig:visualization in testing environment}
\end{figure*}

\subsection{Elimination of Object Attention Bias}

Object attention bias is the phenomenon in which objects with low visibility are ignored. Figure~\ref{fig:proposed unbias attention} (a) shows the average attention of the agent on all objects before and after using the DOA graph for all test floorplans. Without the use of our DOA graph approach, the agent suffers from severe long-tail distribution in the object attention. The average attention difference between the most popular object and the most neglected object is more than tenfold. Objects with high visibility, such as the floor lamp, fridge and sink, dominate the agent's decision-making, while objects with low visibility, such as the cell phone and remote control, cannot play their proper guiding roles. Figure~\ref{fig:proposed unbias attention} (b) shows that although the DOA method makes the average attention for each object to be similar across the entire dataset, there is a correct attention tendency in different scenes.  Our DOA graph-based attention consists of two parts (section 4.2) :  intrinsic attention and adaptive attention. The agent adjusts the two-part attention in different scenes to pay more attention on critical objects. In summary, the proposed DOA graph significantly improves the rationality of attention allocation for most objects, indicating that the improvement in SR, SPL and SAE is indeed from solving the object attention bias problem. We emphasize that DOA graph is a model-agnostic object attention representation method that can be applied to a variety of models and fusion modules.

\subsection{Comparisons to the State-of-the-art}

\begin{table}[t]
\small
\setlength\tabcolsep{4.6pt}
\caption{Comparison with SOTA methods (\%). More experiments are in the appendix D.1.}
\label{tab:SOTA}
\begin{tabular}{c|ccc|ccc}
\hline
\multirow{2}{*}{Method} & \multicolumn{3}{c|}{ALL} & \multicolumn{3}{c}{L\textgreater{}=5} \\
 & SR & SPL & SAE & SR & SPL & SAE \\ \hline \hline
Random & 4.12 & 2,21 & 0.43 & 0.21 & 0.08 & 0.05 \\
SP \cite{yang2018visual} & 62.98 & 38.56 & 24.99 & 52.73 & 33.84 & 23.02 \\
SAVN \cite{wortsman2019learning} & 63.12 & 37.81 & 20.98 & 52.01 & 34.94 & 23.01 \\
ORG \cite{du2020learning} & 67.32 & 37.01 & 25.17 & 58.13 & 35,90 & 27.04 \\
HOZ \cite{zhang2021hierarchical} & 68.53 & 37.50 & 25.98 & 60.27 & 36.61 & 27.68 \\ \hline
Ours (Baseline) & 69.14 & 37.87 & 27.77 & 60.42 & 37.28 & 28.70 \\
\textbf{Ours (DOA+ABED)} & \textbf{74.32} & \textbf{40.27} & \textbf{29.79} & \textbf{67.88} & \textbf{40.36} & \textbf{32.56} \\ \hline
\end{tabular}
\end{table}

As shown in Table~\ref{tab:SOTA}, we compare the test results of our method and other similar methods on the AI2-Thor dataset. All of the random decision navigation indicators are close to 0. Notably, our baseline model outperforms the SOTA method by 0.61/0.15, 0.37/0.67 and 1.97/1.02 in SR, SPL and SAE (ALL/L >= 5, \%). The redundant operations in previous networks aggravate the object attention bias, which explains why the subtraction in our baseline model facilitates learning. Finally, our model with DOA graph-based modules and ABED method outperforms the proposed baseline with the gains of 5.18/7.46, 2.40/3.08 and 2.02/3.86 in SR, SPL and SAE (ALL/L >= 5, \%).

\subsection{Qualitative Analysis}

We visualize the navigation effect of the agent in Figure~\ref{fig:visualization in testing environment}. The direction and stop timing of the rotation are critical, as seen in the trajectories of the success and failure cases. These two decisions are mainly determined by the agent's interpretation of the scene at keyframes when multiple objects can provide rich information. Our DOA graph-based method provides the agent with a more reasonable and unbiased attention allocation at these keyframes, allowing the algorithm to choose the correct rotation direction and stop timing.

\section{Conclusion}

Based on the analysis of the network structure and the failed navigation of previous methods, we identify the agent's object attention bias problem in navigation tasks. To address this problem, we use a directed object attention (DOA) graph, which allows the agent to unbiasedly redistribute object attention. Cross-attention modules (UAIA and UAOA) between the object branch and image branch are designed according to the DOA graph. Our experimental results show that our approach can effectively address the object attention bias problem, greatly improving the agent's navigation ability. Furthermore, we propose an adaptive branch energy distribution (ABED) method for optimizing the aggregation of branch features that performs well in complex models. It is worth noting that we prioritize simplicity in our approach. In future work, we will attempt to determine more concrete relationships between objects so that the agent's navigation can be more clearly interpreted.

\begin{acks}
This paper is supported by the National Natural Science Foundation of China under Grants 61733013, 62173248, 62073245. Suzhou Key Industry Technological Innovation-Core Technology R\&D Program, No. SGC2021035.
\end{acks}

\bibliographystyle{ACM-Reference-Format}
\bibliography{sample-base}


\begin{table*}[!t]
  \caption{Goal-driven navigation datasets. The attributes include: \textbf{Instruction}: type of instruction; \textbf{Observation}: views that the agent can use; \textbf{Scenes}: number of different scenes in the simulation; \textbf{Real-Picture}: pictures taken in real environments; \textbf{Interaction}: whether the agent moves or changes the object state; \textbf{Navigation Graph}: the presumptions of the known environmental topologies; and \textbf{Realistic}: resemblance to real-world environments.}
  \label{tab:dataset}
  \begin{tabular}{|l|c|c|c|c|c|c|c|}
    \hline
    \textbf{Dataset} & \textbf{Instruction} & \textbf{Observation} & \textbf{Scenes} & \textbf{Real-Picture} & \textbf{Interaction} & \textbf{Navigation Graph} & \textbf{Realistic} \\ \hline
    House3D (2018) \cite{wu2018building}    & room name            & single view          & 45622           &                       &                      &                           & weak               \\ \hline
    Habitat (2019) \cite{savva2019habitat}   & object name          & single view          & 90              & \checkmark                   &                      &                           & moderate           \\ \hline
    AI2-Thor (2019) \cite{Kolve2017AI2THORAn} & object name          & single view          & 120             &                       & \checkmark                  &                           & moderate           \\ \hline
    RoboTHOR (2020) \cite{deitke2020robothor}   & object name          & single view          & 89              &                       &                      &                           & strong             \\ \hline
    REVERIE (2020) \cite{qi2020reverie}   & task language        & panoramas            & 90              & \checkmark                   &                      & \checkmark                       & weak               \\ \hline
    FAO (2021) \cite{zhu2021soon}       & locate language      & panoramas            & 90              & \checkmark                   &                      & \checkmark                       & weak               \\ \hline
    \end{tabular}
\end{table*}

\appendix
\section{More Analyses on Datasets}

Table~\ref{tab:dataset} summarizes the most commonly used datasets for object navigation tasks. We compare these datasets in terms of the instructions given, the views used, the variety and realism of the scenes, the ability of the agent to interact, the presence or absence of a predefined navigation map, and the ability to map the environment to reality. Since our method is based on the input of object names and single views, in addition to using the AI2-Thor dataset discussed in the main text, we also conduct experiments on the RoboTHOR dataset.

\section{Object Attention Bias}

\begin{figure}[ht]
    \centering
    \includegraphics[width=0.5\textwidth]{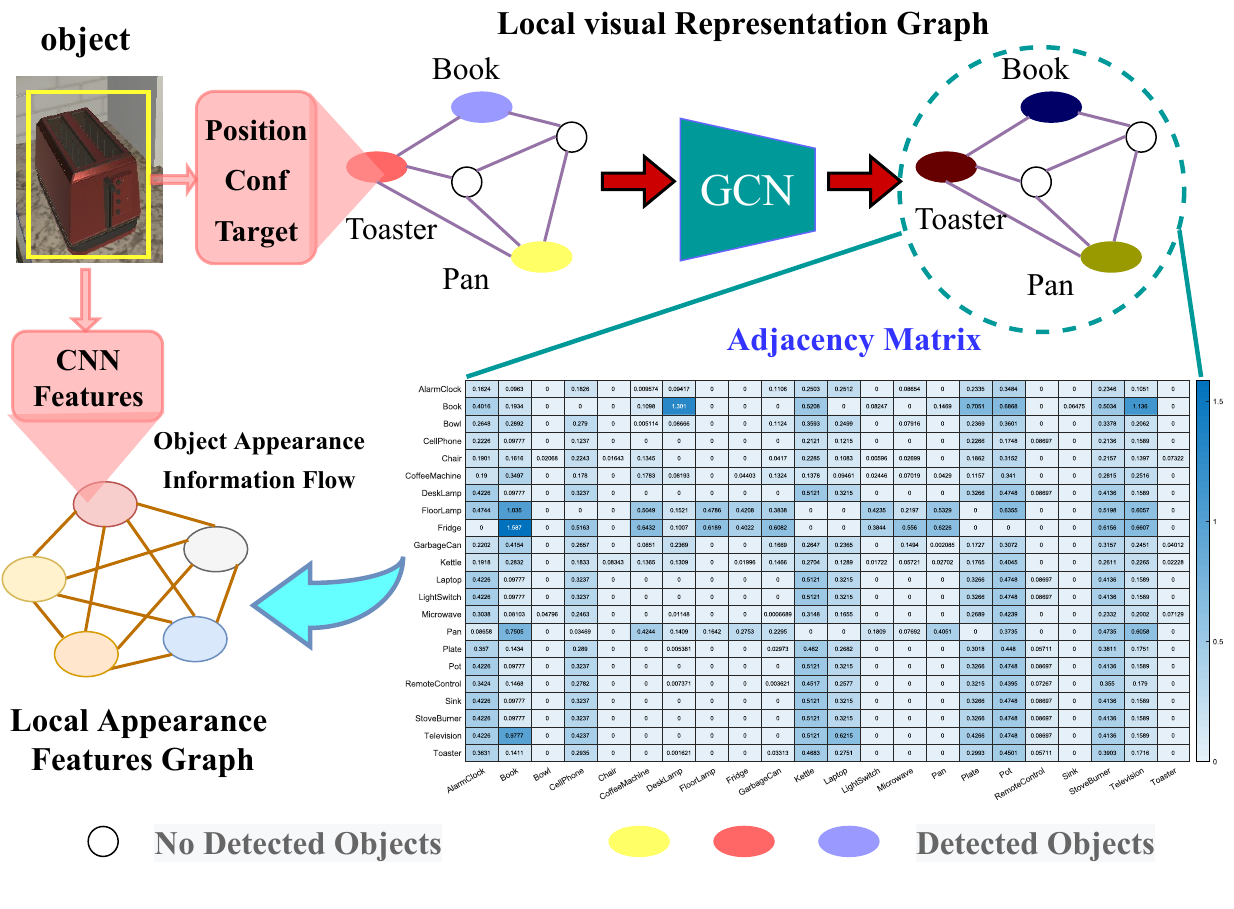}
    \caption{The object feature embedding method commonly used in previous works \cite{du2020learning,zhang2021hierarchical}. A local visual representation graph (containing the position, confidence, and target) is embedded in the adjacency matrix with a GCN. The adjacency matrix is used to fuse object features with strong correlations in the local appearance features graph.}
    \label{fig:object GCN problem}
\end{figure}

\begin{figure*}[ht]
    \centering
    \includegraphics[width=\textwidth]{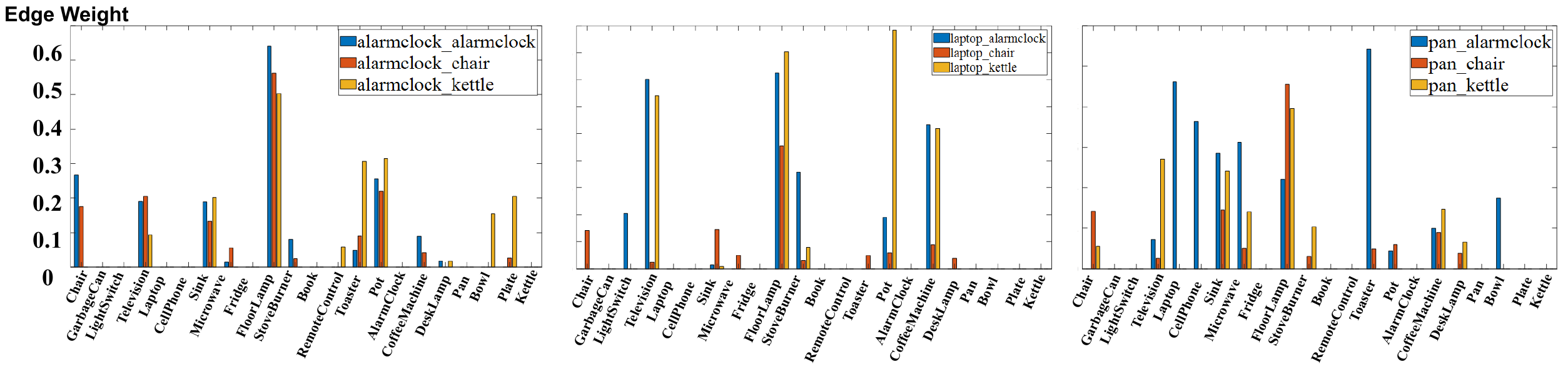}
    \caption{Visualization of object attention bias in object GCN. A\_B represents the weight of object B and all 22 objects in the adjacency matrix while searching for target A.}
    \label{fig:GCN bias attention}
\end{figure*}

\begin{figure}[ht]
    \centering
    \includegraphics[width=0.48\textwidth]{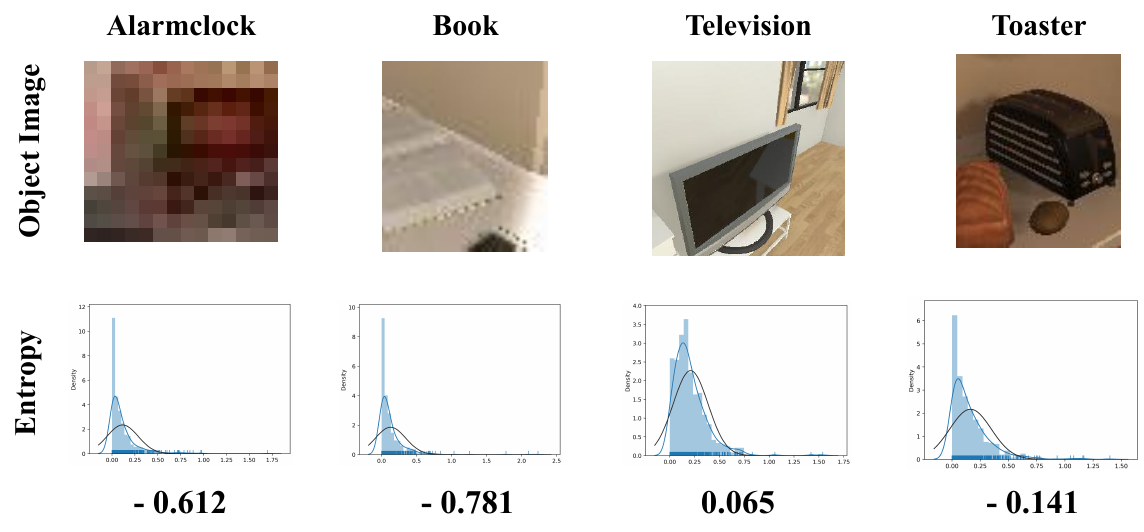}
    \caption{The image and entropy of different objects in different environments.}
    \label{fig:object entropy}
\end{figure}

\begin{table*}[]
\setlength\tabcolsep{7.2pt}
\caption{Comparison of introducing object semantics and position into the GCN node features (\%).}
\label{tab:object semantic}
\begin{tabular}{ccc|ccc|ccc|l}
\hline
\multirow{2}{*}{GCN} & \multirow{2}{*}{Semantics} & \multirow{2}{*}{Position} & \multicolumn{3}{c|}{ALL} & \multicolumn{3}{c|}{L\textgreater{}=5} & \multirow{2}{*}{Episode Length} \\
                     &        &                           & SR     & SPL    & SAE    & SR          & SPL         & SAE        &                                 \\ \hline \hline
\checkmark                  &                 &                 & $68.53_{\pm1.32}$  & $37.50_{\pm1.01}$  & $25.98_{\pm0.96}$  & $60.27_{\pm1.67}$       & $36.61_{\pm1.12}$       & $27.68_{\pm1.31}$      & \multicolumn{1}{c}{$27.25_{\pm0.87}$}       \\
\checkmark                  & \checkmark              &               & $69.45_{\pm0.76}$  & $37.36_{\pm0.99}$  & $27.65_{\pm1.31}$  & $61.02_{\pm1.09}$       & $37.15_{\pm0.32}$       & $29.30_{\pm1.78}$      & \multicolumn{1}{c}{$26.02_{\pm0.86}$}       \\ 
\checkmark                  &              &   \checkmark             & $69.12_{\pm0.90}$  & $38.19_{\pm0.56}$  & $27.23_{\pm0.88}$  & $60.76_{\pm1.78}$       & $37.94_{\pm1.12}$       & $29.01_{\pm1.34}$      & \multicolumn{1}{c}{$25.33_{\pm1.24}$}       \\\hline
\end{tabular}
\end{table*}

We discover the object attention bias problem during our research on object features GCN, as discussed in \cite{du2020learning, zhang2021hierarchical}. In the main text, we briefly analyzed this problem. Here, we elaborate on why GCNs have issues with object attention bias. As shown in Figure~\ref{fig:object GCN problem}, the object GCN is divided into two stages: (\romannumeral1) Generating the adjacency matrix, and (\romannumeral2) Transferring information between objects using the adjacency matrix. The object attention bias problem arises in both stages.

\subsection{Bias in Generating the Adjacency Matrix}

The adjacency matrix is generated based on the relationships between the bounding boxes of objects, which defines the correlation between objects. However, after training, the network determines the importance of an object based on the size of its bounding box and its confidence score, thereby ignoring the category and position of the object, which are both more important features. In Table~\ref{tab:object semantic}, we explicitly add the object semantics and position to the GCN node features of the HOZ \cite{zhang2021hierarchical} method and find that the agent's navigation performance improves slightly. The results show that it is important for the agent to pay more attention to what the object is and where the objects are; however, this feature introduction cannot effectively resolve the object attention bias generated during training the GCN. Furthermore, since the agent usually can view only a few objects at a time, the object features graph is sparse and thus more susceptible to bias. Figure~\ref{fig:GCN bias attention} shows the weights on the edge connecting some specified objects and other objects in the adjacency matrix when searching for different targets in the same bedroom scene. Due to the sparsity of the objects, most edge weights are 0. It is worth noting that regardless of what object is being searched for, the floor lamp has a high connection weight with the other objects. This illustrates that the volume and clarity of the floor lamp cause the agent to overestimate its importance in the bedroom scene.

\subsection{Bias in Convolutional Object Visual Features}

The convolution of object visual features with a biased adjacency matrix clearly leads to biased results. However, in this section, we discuss that even if the object bias in the adjacency matrix is not considered, the convolution of the object visual features can also generate this bias. Figure~\ref{fig:object entropy} shows the original images of the objects and their information entropy calculated by the k-nearest neighbor (KNN) method \cite{kozachenko1987sample}. As a result, even if an object is recognized, it is likely to be ignored due to the disadvantage of its feature information richness.

\section{More Ablation Experiments}

\begin{table*}[]
\setlength\tabcolsep{6.1pt}
\caption{Navigation performance after replacing some modules in the model (\%). Our Best Model represents the model used in our main text. }
\label{tab:more ablation}
\begin{tabular}{cc|ccc|ccc|c}
\hline
\multicolumn{2}{c|}{\multirow{2}{*}{Module Replacement}}                  & \multicolumn{3}{c|}{ALL} & \multicolumn{3}{c|}{L\textgreater{}=5} & \multirow{2}{*}{Episode Length} \\
\multicolumn{2}{c|}{}                                                 & SR     & SPL    & SAE    & SR          & SPL         & SAE        &                                 \\ \hline \hline
\multicolumn{2}{c|}{Our Best Model}                                   & $74.32_{\pm0.89}$  & $40.27_{\pm1.01}$  & $29.79_{\pm0.80}$  & $67.88_{\pm1.05}$       & $40.36_{\pm1.23}$       & $32.56_{\pm1.76}$      & $22.86_{\pm1.32}$                           \\ \hline
\multicolumn{1}{c|}{\multirow{2}{*}{Backbone}}         & ResNet50     & $74.12_{\pm1.23}$  & $35.98_{\pm1.56}$  & $28.76_{\pm0.90}$  & $66.77_{\pm1.12}$       & $33.50_{\pm1.27}$       & $32.34_{\pm1.52}$      & $27.96_{\pm0.88}$                           \\
\multicolumn{1}{c|}{}                                  & ResNet101    & $70.21_{\pm1.45}$  & $35.56_{\pm1.07}$  & $26.77_{\pm1.34}$  & $65.43_{\pm0.81}$       & $31.45_{\pm1.23}$       & $29.70_{\pm0.99}$      & $28.21_{\pm1.29}$                           \\ \hline
\multicolumn{1}{c|}{Object Detector}                   & Ground Truth & $76.12_{\pm1.48}$  & $59.12_{\pm1.10}$  & $30.98_{\pm1.27}$  & $69.07_{\pm0.65}$       & $54.93_{\pm1.35}$       & $35.15_{\pm0.93}$      & $17.45_{\pm1.05}$                           \\ \hline
\multicolumn{1}{c|}{\multirow{3}{*}{Pixel Index Emb.}} & No           & $74.54_{\pm1.20}$  & $38.14_{\pm0.82}$  & $29.13_{\pm1.03}$  & $67.76_{\pm1.38}$       & $38.55_{\pm1.07}$       & $32.67_{\pm1.21}$      & $24.21_{\pm0.94}$                           \\
\multicolumn{1}{c|}{}                                  & 2-D          & $74.29_{\pm1.22}$  & $40.54_{\pm0.79}$  & $29.38_{\pm1.62}$  & $67.81_{\pm1.09}$       & $40.35_{\pm1.67}$       & $32.29_{\pm1.24}$      & $22.59_{\pm0.93}$                           \\
\multicolumn{1}{c|}{}                                  & Relative     & $74.10_{\pm0.85}$  & $40.34_{\pm1.07}$  & $29.33_{\pm1.55}$  & $67.45_{\pm1.38}$       & $40.19_{\pm1.43}$       & $31.98_{\pm1.33}$      & $23.45_{\pm0.74}$                           \\ \hline
\multicolumn{1}{c|}{DOA Graph}                         & Undirected   & $73.15_{\pm1.17}$  & $38.44_{\pm1.26}$  & $27.43_{\pm1.63}$  & $66.33_{\pm1.44}$       & $37.60_{\pm1.10}$       & $30.95_{\pm1.28}$      & $26.84_{\pm1.36}$                           \\ \hline
\end{tabular}
\end{table*}

\begin{figure}[t]
    \centering
    \includegraphics[width=0.45\textwidth]{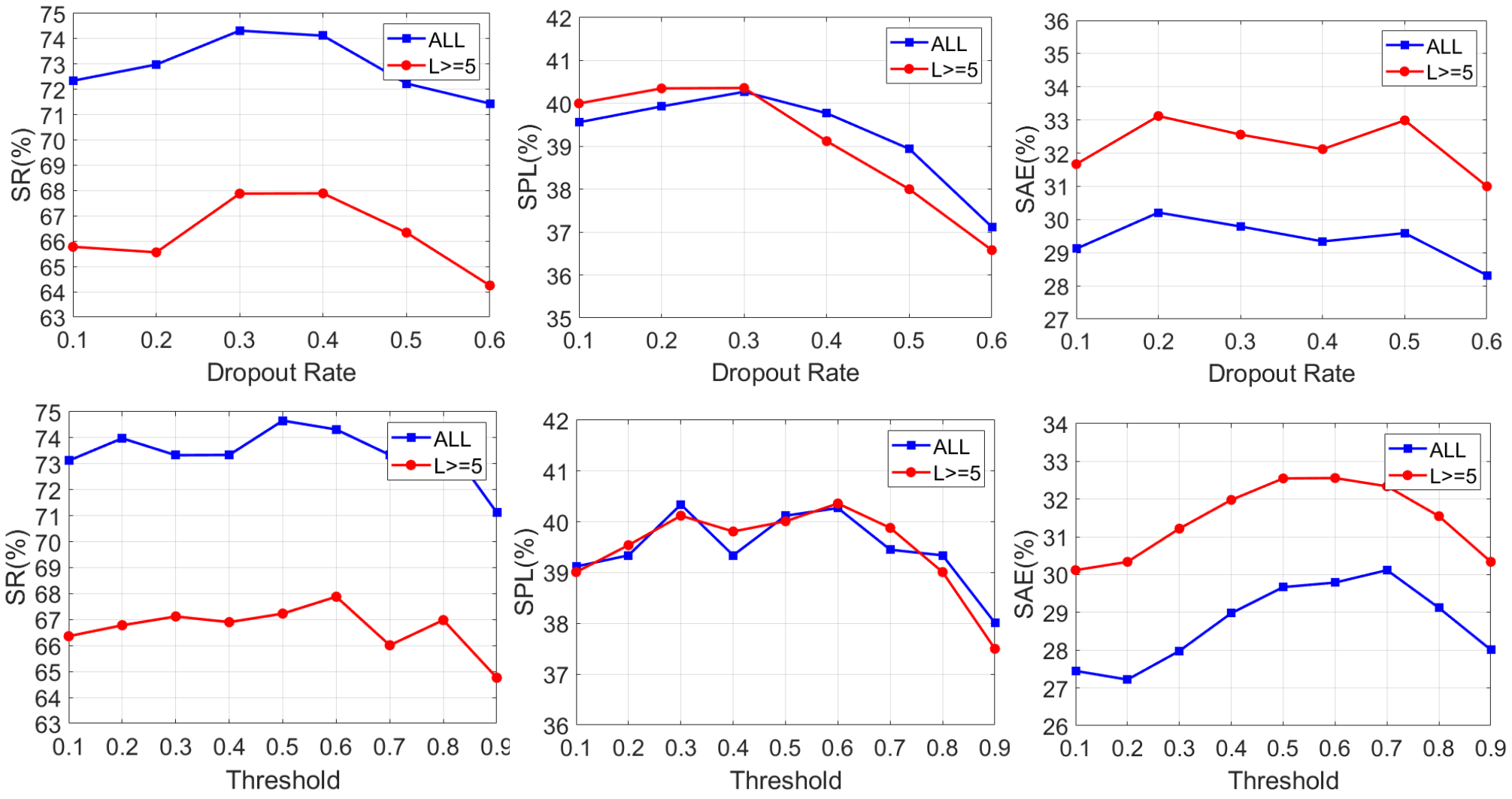}
    \caption{Experiments to explore the most reasonable hyperparameters, such as the dropout rate and the threshold of confidence filter.}
    \label{fig:hyperparameters}
\end{figure}

\subsection{Choice of Backbone}

In the main text, we use ResNet18 as the backbone network for extracting the global image features. Additionally, we also experiment with backbones of different depths. Table~\ref{tab:more ablation} shows the results after we replace the backbone with ResNet50 and ResNet101. Compared to ResNet18, the use ResNet50 decreases the SR, SPL and SAE by 0.20/1.11, 4.29/6.86 and 1.03/0.22 (ALL/L>=5,\%). The losses in the SR, SPL and SAE when using ResNet101 as the backbone network are even greater. It is clear that the SPL and the episode length are most severely affected. The explanation may be that excessive high-level global image information degrades local information, thereby weakening the effect of cross-attention with the object branch and impairing the action selection at each step in the overall path. As a result, the navigation distance is greatly increased, which affects the SPL.

\subsection{Importance of Accurate Object Information}

In our model, we apply Faster-RCNN \cite{ ren2015faster}, which was trained on 50\% \cite{zhang2021hierarchical} of the training data from the AI2-Thor dataset, to extract the object information. Table~\ref{tab:more ablation} shows the results when the ground truth object information is used as the input of the object branch, improving the SR, SPL and SAE by 1.80/1.19, 18.85/14.57 and 1.19/2.59 (ALL/L >=5, \%). Similar to the conclusion presented in \cite{zhang2021hierarchical}, the ground truth object information has a strong  effect on the optimization of navigation path. Notably, our DOA graph method significantly increases the upper bound of the model, and when more accurate and efficient features are used, our model performs better.

\begin{figure*}[ht]
    \centering
    \includegraphics[width=\textwidth]{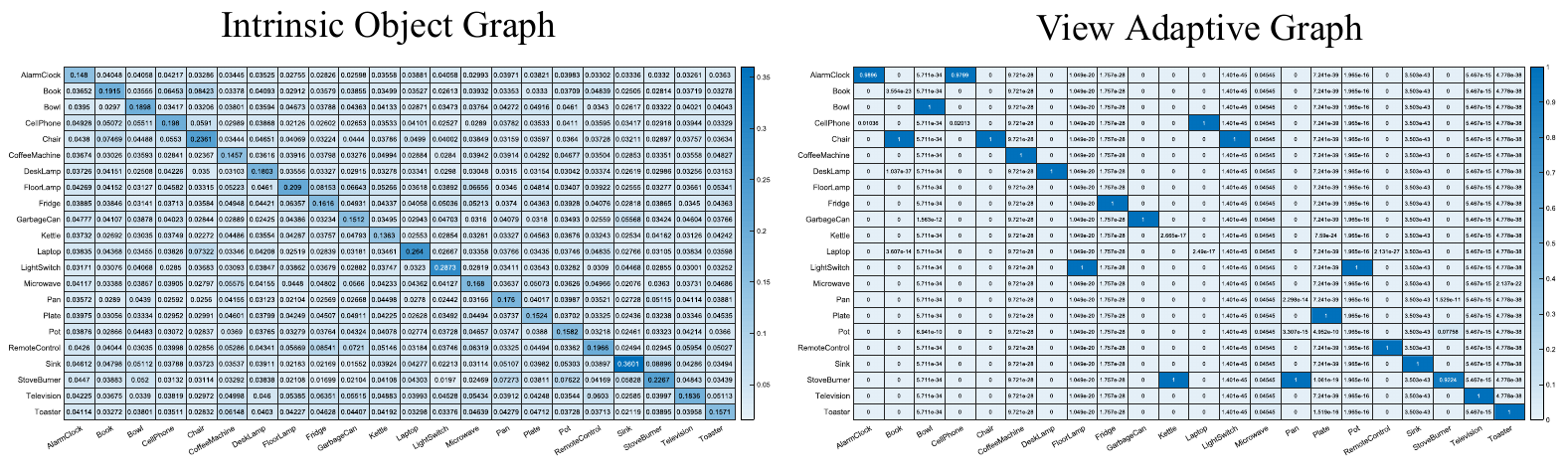}
    \caption{Learned adjacency matrices of the intrinsic object graph and view adaptive graph.}
    \label{fig:DOA visual}
\end{figure*}

\begin{table*}[ht]
\setlength\tabcolsep{7.2pt}
\large
\caption{Comparision with SOTA methods in RoboTHOR \cite{deitke2020robothor} (\%)}
\label{tab:compare in RoboTHOR}
\begin{tabular}{c|ccc|ccc|c}
\hline
\multirow{2}{*}{Method}  & \multicolumn{3}{c|}{ALL}                         & \multicolumn{3}{c|}{L\textgreater{}=5} & \multirow{2}{*}{Episode Length}           \\
                         & SR             & SPL            & SAE            & SR             & SPL            & SAE &           \\ \hline \hline
Random                   & $0.00_{\pm0.00}$           & $0.00_{\pm0.00}$           & $0.00_{\pm0.00}$           & $0.00_{\pm0.00}$           & $0.00_{\pm0.00}$           & $0.00_{\pm0.00}$     &   $3.01_{\pm0.32}$        \\
LSTM+A3C                 & $21.35_{\pm0.23}$          & $9.28_{\pm0.54}$           & $7.03_{\pm0.34}$           & $11.21_{\pm0.41}$          & $7.65_{\pm0.26}$           & $6.33_{\pm0.86}$      &   $87.29_{\pm1.95}$       \\
SP \cite{yang2018visual}                      & $27.43_{\pm0.60}$          & $17.49_{\pm0.36}$          & $15.21_{\pm0.33}$          & $20.98_{\pm0.97}$          & $16.03_{\pm0.48}$          & $13.99_{\pm0.44}$      &   $68.18_{\pm1.27}$    \\
SAVN \cite{wortsman2019learning}                    & $28.97_{\pm0.84}$          & $16.59_{\pm0.65}$          & $13.21_{\pm0.37}$          & $22.89_{\pm0.98}$          & $15.21_{\pm0.41}$          & $13.77_{\pm0.30}$     &   $67.22_{\pm0.93}$      \\
ORG \cite{du2020learning}                    & $30.51_{\pm0.21}$          & $18.62_{\pm0.19}$          & $14.71_{\pm0.47}$         & $23.89_{\pm0.55}$          & $14.91_{\pm0.38}$          & $13.73_{\pm0.50}$       &   $69.17_{\pm2.13}$   \\
HOZ \cite{zhang2021hierarchical}                    & $31.67_{\pm0.87}$          & $19.02_{\pm0.49}$          & $15.44_{\pm0.63}$          & $24.32_{\pm0.48}$          & $14.81_{\pm0.23}$          & $14.22_{\pm0.88}$       &   $66.26_{\pm1.88}$   \\ \hline
Ours (Baseline)          & $31.92_{\pm0.37}$          & $18.88_{\pm0.18}$          & $16.37_{\pm0.33}$          & $24.72_{\pm0.26}$          & $15.56_{\pm0.35}$          & $15.18_{\pm0.61}$        &   $65.31_{\pm1.69}$  \\
\textbf{Ours (DOA+ABED)} & $\textbf{36.22}_{\pm0.36}$ & $\textbf{22.12}_{\pm0.42}$ & $\textbf{18.70}_{\pm0.59}$ & $\textbf{30.16}_{\pm0.28}$ & $\textbf{18.32}_{\pm0.34}$ & $\textbf{19.21}_{\pm0.52}$  &  $\textbf{61.24}_{\pm1.23}$\\ \hline
\end{tabular}
\end{table*}

\begin{figure*}[ht]
    \centering
    \includegraphics[width=0.9\textwidth]{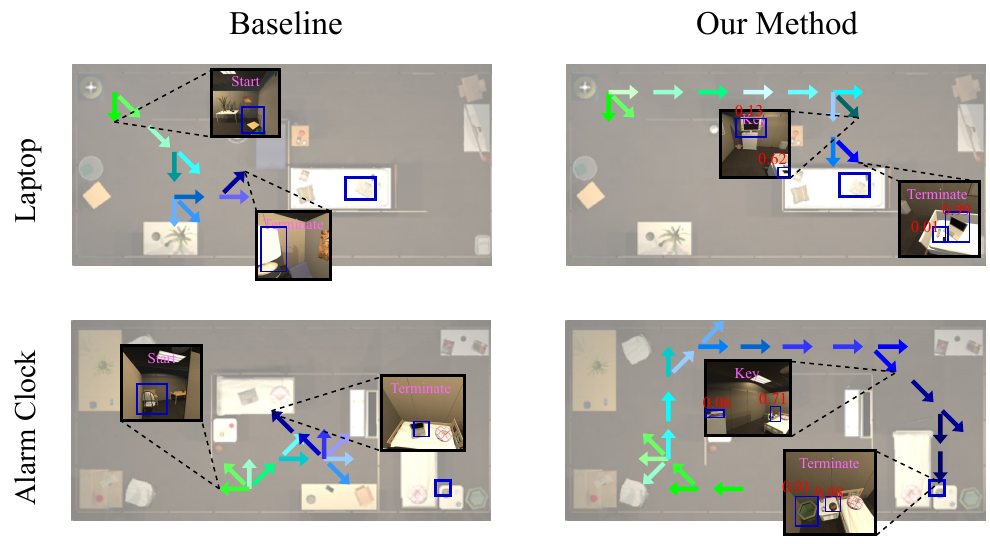}
    \caption{Visualization of the navigation trajectory with the RoboTHOR dataset. The trajectory of the agent is indicated by the green and blue arrows, with the green indicating the beginning and the blue indicating the end of the trajectory. The red value in the object detection box represents the attention weight of the object.}
    \label{fig:visual in RoboTHOR}
\end{figure*}

\begin{figure}[ht]
    \centering
    \includegraphics[width=0.45\textwidth]{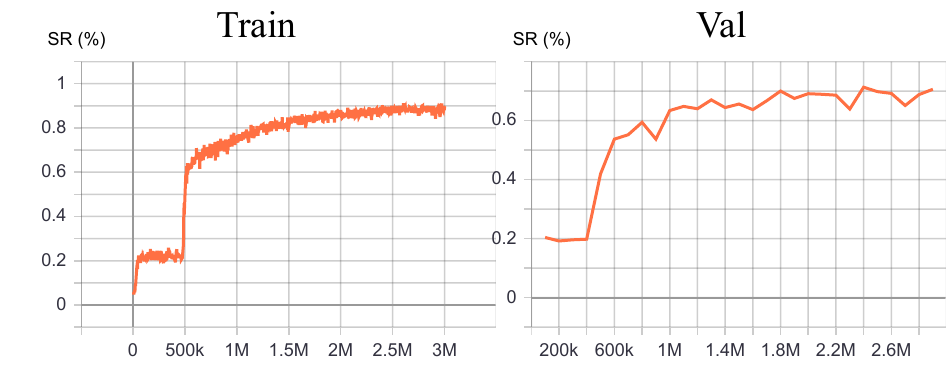}
    \caption{The performance of the agent on the training and validation  datasets during training. The agent is tested on the validation  dataset every $10^5$ navigation episodes.}
    \label{fig:training process}
\end{figure}

\subsection{Pixel Index Embedding}

In the UAIA module, the pixel index embeddings provide each pixel’s position token, which is important in the muti-head attention mechanism. In Table~\ref{tab:more ablation}, we compare the results of the following encodings: 
\begin{itemize}[leftmargin=*]
  \item [1)] 
  Providing no positional information (No): Considering the image feature maps as a bag of pixel vectors.      
  \item [2)]
  1-dimensional pixel index embedding (Our Best Model): Considering the image feature maps as a sequence of pixel vectors in order.
  \item [3)]
  2-dimensional pixel index embedding (2-D): Considering the image feature maps as a grid of pixel vectors in two dimensions.
  \item [4)]
  Relative pixel index embedding (Relative): Considering the relative distance between pixels rather than their absolute position to encode the spatial information.
\end{itemize}
The results show that without pixel index embeddings, the SPL is reduced by approximately 2\%. However, when using 1-D, 2-D or relative pixel index embeddings, the effect is essentially the same. We speculate that this occurs because we perform position embeddings on feature maps with smaller spatial dimensions instead of on the original image, which has larger spatial dimensions. Hence, the process of learning the spatial relations at this resolution is equally easy for the different positional encoding strategies.

\subsection{Directivity in DOA Graph }

In the main text, we repeatedly emphasize that our proposed DOA graph is a directed graph; thus, the correlation between object B and object A when searching for object A is different from the correlation between object A and object B when searching for object B. Table~\ref{tab:more ablation} illustrates that the navigation improvement decreases significantly when the DOA graph is an undirected graph. Fundamentally, the directed object graph follows the relative relationships between objects, while the undirected object graph follows the absolute relationships between objects. The absolute object relationships reduce the adaptability of the agent in modeling object relationships in different scenes, thus reducing its performance in unseen environments.

\subsection{Hyperparameters}

The hyperparameters are also an important factor in determining the final navigation performance of the agent. However, due to the complexity of the object navigation task, small changes in the network structure have a considerable impact on the selection of the hyperparameters. Therefore, as shown in Figure~\ref{fig:hyperparameters}, we conduct extensive experiments on the final model with different dropout rates and confidence filter thresholds. The experimental results show that when a hyperparameter changes, the change rules of the three metrics (the SR, SPL and SAE) differ. Thus, we consider all three metrics when choosing the most appropriate hyperparameters for our model.

\section{More comparisons with the related works}

\subsection{Comparisons in RoboTHOR Dataset}

We compare our method with the other related works on the AI2-Thor dataset in the main text. We consider that the scenes in AI2-Thor are relatively simple, so the RoboTHOR \cite{deitke2020robothor} dataset is used to verify the effectiveness of our method in complex environment. RoboTHOR has public 75 apartments, we choose 60 for training, 5 for validation and 10 for testing. 

Table~\ref{tab:compare in RoboTHOR} illustrates the performance of our method and other related methods on the RoboTHOR. Obviously, the various metrics on the RoboTHOR are generally low, which indicates that object navigation on RoboTHOR is much more difficult than AI2-Thor. Moreover, The episode length of navigation on RoboTHOR is 3 times that of AI2-Thor, which indicates that each navigation scene in RoboTHOR is much more complicated. Nevertheless, out method still outperforms the proposed baseline with an obvious margin by 4.3/5.44, 3.24/2.76 and 2.33/4,03 in SR, SPL and SAE (ALL/L>=5, \%).

\section{Qualitative Results}

\subsection{The DOA Graph Visualization}

Our proposed DOA graph has two parts: the intrinsic object graph, which represents the ubiquitous inherent relationship between the objects, and the view adaptive graph, which changes adaptively according to the input at each timestep. A weighted sum of the two graphs gives the attention score assigned to each object at this timestep. Figure~\ref{fig:DOA visual} shows the adjacency matrix of the two graphs. For the adjacency matrix of the intrinsic object graph, the largest value is on the diagonal, indicating that the self-connecting edge has the largest weight. This result is consistent with our assumption that when the target object is identified, most of the attention is focused on the target object. For the adjacency matrix of the view adaptive graph, we randomly select the view adaptive attention when looking for different targets as a column. Interestingly, at each timestamp in the view adaptive graph, only one object's attention weight is close to 1, while the other object's attention weight is close to 0. We speculate that because there are often very few objects observed at each timestamp and most of these objects are essentially irrelevant to the target object, the view adaptive graph directs all attention to the most important object. In the weighted sum of the two graphs, the weight of the intrinsic object graph is 0.95, while the weight of the view adaptive graph is only 0.05. Therefore, although the attention distribution of the view adaptive graph is extreme, it does not affect the robustness of the overall attention distribution. In contrast, the attention concentration in the view adaptive graph reasonably increases the weight of important objects in view.

\subsection{Training Process}

Figure~\ref{fig:training process} shows the agent's navigation success rate (SR) on the training and validation datasets during reinforcement learning. We observe that after the agent has learned 0.5 M episodes, the navigation success rate greatly improved, and the SR curve converged after 2.5 M episodes. Due to computational resource constraints, we train our models on 3 M navigation episodes in our usual experiments. To obtain the best model of the current network structure, we can increase the number of training episodes to 6 M.

\subsection{Navigation Trajectory in RoboTHOR}

Since the AI2-Thor dataset only includes simple room constructions, we also use the RoboTHOR dataset, which includes more complex indoor environments, to verify the effectiveness of our model. As shown in Figure~\ref{fig:visual in RoboTHOR}, in the RoboTHOR environment, the agent spends more time searching for the target object. Compared with the baseline, the agent can approach the target step-by-step in a continuous exploration with our proposed DOA graph.

\end{document}